\documentclass[runningheads]{llncs}
\usepackage[T1]{fontenc}
\usepackage{graphicx}
\usepackage{marvosym}
\usepackage{amsmath}
\usepackage{amsfonts,amssymb}
\usepackage{booktabs}
\usepackage{hyperref}
\usepackage{cite}

\hypersetup{
    colorlinks=true, 
    linkcolor=blue,  
    citecolor=blue,  
    urlcolor=blue    
}

\begin{document}

\title{ArtNeRF: A Stylized Neural Field for 3D-Aware Cartoonized Face Synthesis}

\author{Zichen Tang\inst{1}\and Hongyu Yang\inst{2}\textsuperscript{(\Letter)}}

\institute{Institute of Artificial Intelligence, Beihang University, Beijing, China \\ \email{\{zctang, hongyuyang\}@buaa.edu.cn}}

\maketitle

\begin{abstract}
Recent advances in generative visual models and neural radiance fields have greatly boosted 3D-aware image synthesis and stylization tasks. However, previous NeRF-based work is limited to single scene stylization, training a model to generate 3D-aware cartoon faces with arbitrary styles remains unsolved. We propose ArtNeRF, a novel face stylization framework derived from 3D-aware GAN to tackle this problem. In this framework, we utilize an expressive generator to synthesize stylized faces and a triple-branch discriminator module to improve the visual quality and style consistency of the generated faces. Specifically, a style encoder based on contrastive learning is leveraged to extract robust low-dimensional embeddings of style images, empowering the generator with the knowledge of various styles. To smooth the training process of cross-domain transfer learning, we propose an adaptive style blending module which helps inject style information and allows users to freely tune the level of stylization. We further introduce a neural rendering module to achieve efficient real-time rendering of images with higher resolutions. Extensive experiments demonstrate that ArtNeRF is versatile in generating high-quality 3D-aware cartoon faces with arbitrary styles. Our source codes are publicly available at \href{https://github.com/silence-tang/ArtNeRF}{https://github.com/silence-tang/ArtNeRF}.

\end{abstract}

\section{Introduction}
With the rise of concepts like Metaverse and Artificial Intelligence Generated Content (AIGC), 3D stylization technology has become increasingly pivotal in various application scenarios such as AR/VR. In this work, we address a novel task of 3D-aware image stylization: given a latent identity code, a style image, and multiple camera poses, the model should generate 3D-aware stylized faces with high multi-view consistency while preserving the style characteristics of the style image. The challenges of this task are primarily threefold: (1) How to ensure the style consistency between the style image and the generated image. (2) How to prevent structural information such as the pose of the reference style image from leaking into the generated image. (3) How to guarantee high multi-view consistency and visual quality of the results while achieving efficient real-time rendering.\label{intro:}

\begin{figure}
\includegraphics[width=\textwidth]{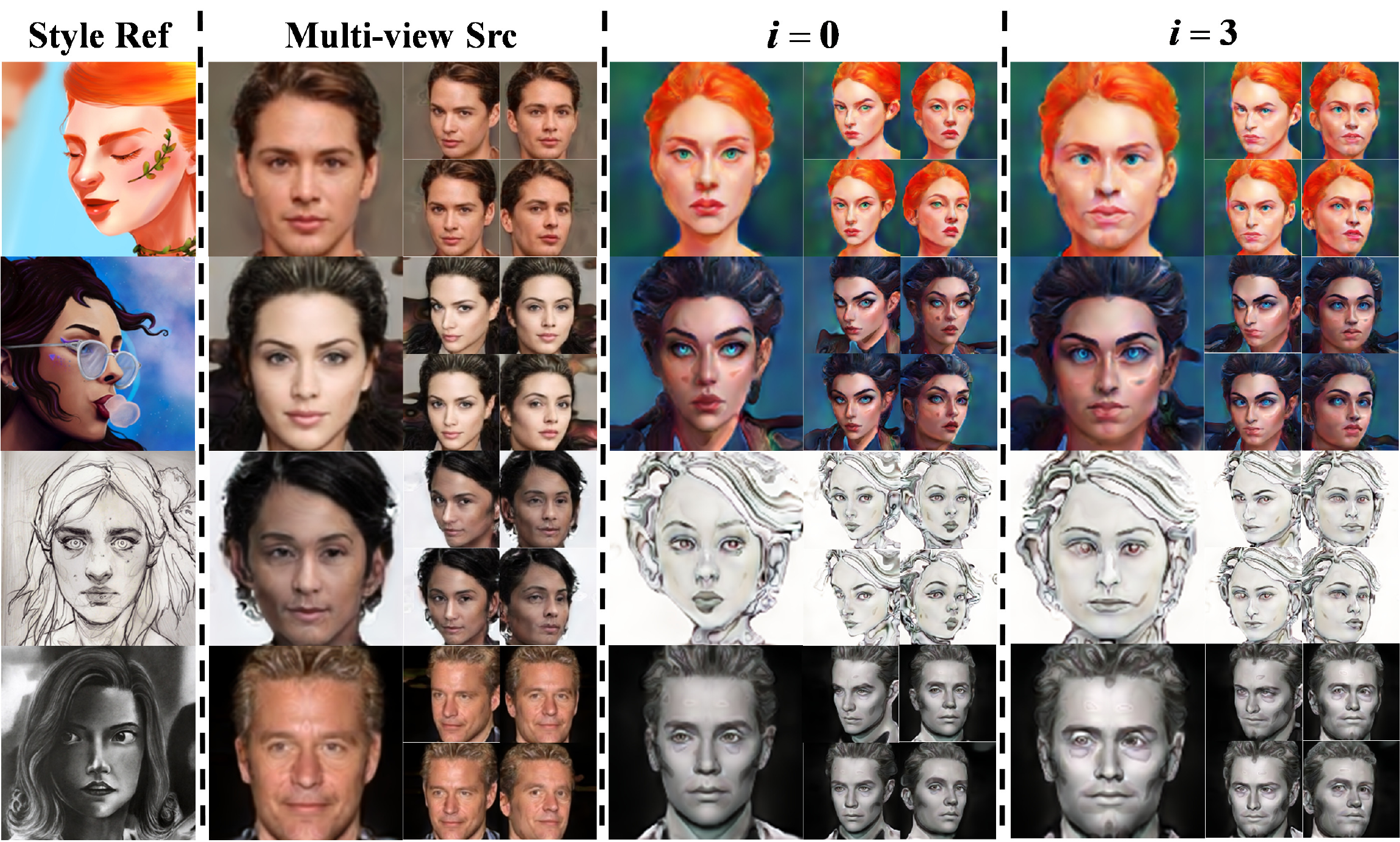}
\caption{Multi-view 3D-aware faces with arbitrary styles generated by our model. We evenly select 5 views within a reasonable range where $\rm{pitch}\in[\frac{\pi}{2}-0.2, \frac{\pi}{2}+0.2]$ and $\rm{yaw}\in[\frac{\pi}{2}-0.4, \frac{\pi}{2}+0.4]$.} \label{fig:1}
\vspace{-10pt}
\end{figure}

Many existing 2D methods \cite{ref_proc1,ref_proc2,ref_proc3} encode source and reference images into content and attribution latent codes, then combine these codes for reconstruction or style transfer. While these methods support arbitrary reference images, they often yield results with low visual quality and inconsistent style. Additionally, none of these 2D methods can generate 3D-aware results. More recently, \cite{ref_proc16} proposes a domain-adaption framework and \cite{ref_proc17} presents a novel stylefield for 3D-aware image stylization. Although these methods achieve high visual quality in 3D-aware results, they are limited to fixed styles and face challenges like high GPU memory consumption during training owing to their 3D representations.

To overcome the limitations of existing methods, we propose \textbf{ArtNeRF}, a novel 3D-aware GAN framework for generating multi-view faces with arbitrary styles from given reference images. ArtNeRF features an expressive 3D-aware generator paired with a triple-branch discriminator, achieving rapid and high-quality face stylization. As for the generator, we design better implementation practices to reliably enhance the generation quality of pi-GAN \cite{ref_proc11}. Specifically, we design dense skip connection layers in the original backbone to strengthen the reuse of feature maps from different semantic layers and discard the progressive growing training strategy. We then design a neural rendering module comprising $1\times1$ convolutional layers and a skip connection mechanism for image super-resolution and faster rendering. For style representation extraction, we leverage a style encoder based on contrastive learning, effectively mapping reference images to low-dimensional style codes. To address training instability stemming from cross-domain learning discrepancies, we propose an adaptive style blending module that dynamically adjust the blending ratio of style control vectors to ensure a smooth training process. Finally, our triple-branch discriminator module consists of three discriminators with analogous architectures. The first two aid the generator in synthesizing faces adhering to the distributions of the source and target domains, while the third one with an embedder head is capable of improving style consistency between synthesized faces and the reference images. Our contributions can be summarized as follows: \\
\indent - We propose a novel 3D-aware image arbitrary stylization task, where the synthesized results should emulate the style characteristics of the style reference image while maintaining strong multi-view consistency. Correspondingly, We design ArtNeRF, a framework based on 3D-aware GAN to realize this goal.\\
\indent - We introduce a self-adaptive style blending module to inject style information into the generator and a triple-branch discriminator to guarantee style consistency. Incorporated with our two-stage training strategy, the cross-domain adaption process can be smoothed and stabilized effectively. \\
\indent - By designing dense skip connections between sequential layers and incorporating a neural rendering module, we boost the generator backbone of pi-GAN, leading to efficient real-time rendering and better visual quality. \\

\section{Related works}
\subsection{Style Transfer with 2D GAN}
MUNIT~\cite{ref_proc1}, FUNIT~\cite{ref_proc2}, DRIT++~\cite{ref_proc3} and StarGANv2~\cite{ref_proc4} are seminal works that focus on reference-guided image style transfer using GAN~\cite{ref_proc5}. Subsequently, several methods have been proposed to achieve style transfer in specific style domains. CartoonGAN~\cite{ref_proc6} introduces various losses suitable for general photo cartoonization while ChipGAN~\cite{ref_proc7} utilizes an adversarial loss for Chinese ink painting style transfer with constraints on strokes and ink tone. Some recent works combine expressive backbones with unique designs, further boosting the artistic effects of synthesized images. BlendGAN~\cite{ref_proc8} proposes a style encoder and employs a style-conditioned discriminator to generate 2D faces with arbitrary styles. Pastiche Master~\cite{ref_proc9} employs a dual-path style generation network and introduces multi-stage fine-tuning strategies, achieving facial cartoonization with fixed styles. However, none of these methods can generate vivid 3D-aware results.

\subsection{3D-aware Image Synthesis}
In the realm of 3D-aware image synthesis, we mainly focus on NeRF-based methods. GRAF~\cite{ref_proc10}, pi-GAN~\cite{ref_proc11} and GIRAFFE~\cite{ref_proc12} combine GAN with NeRF~\cite{ref_proc13} to learn a 3D representation from 2D images, thereby enabling novel view synthesis. Other endeavors aim to narrow the gap in visual quality between 3D models and 2D GAN models. For instance, GRAM~\cite{ref_proc14} introduces an implicit neural representation based on learnable 2D manifolds, enhancing the quality of synthesized images with reduced sampling points. EG3D~\cite{ref_proc15} presents an effective tri-plane representation for high-quality 3D-aware image synthesis. More recently, some 3d-aware stylization works like 3DAvatarGAN~\cite{ref_proc16} and DeformToon3D~\cite{ref_proc17} are proposed to generate 3D-aware avatars with specific styles. Nevertheless, these methods incur high training costs and are limited in their ability to handle arbitrary styles with a single trained model.

\section{Method}
Given an identity code $\boldsymbol{z}$, a reference style image $\boldsymbol{X_s}$ and camera poses $\boldsymbol{p}$, we aim to generate high-quality 3D-aware stylized faces which are supposed to maintain consistent across various views. We firstly give preliminaries in Sec \ref{sec:3.1}. To solve the challenges discussed in the introduction, we leverage a style encoder to extract style embeddings of reference images in Sec \ref{sec:3.2}, a novel generative radiance field to achieve efficient style blending and rendering in Sec \ref{sec:3.3}, and a triple-branch discriminator network to supervise the 3D-aware generator and enhance style consistency in Sec \ref{sec:3.4}.

\begin{figure}
\includegraphics[width=\textwidth]{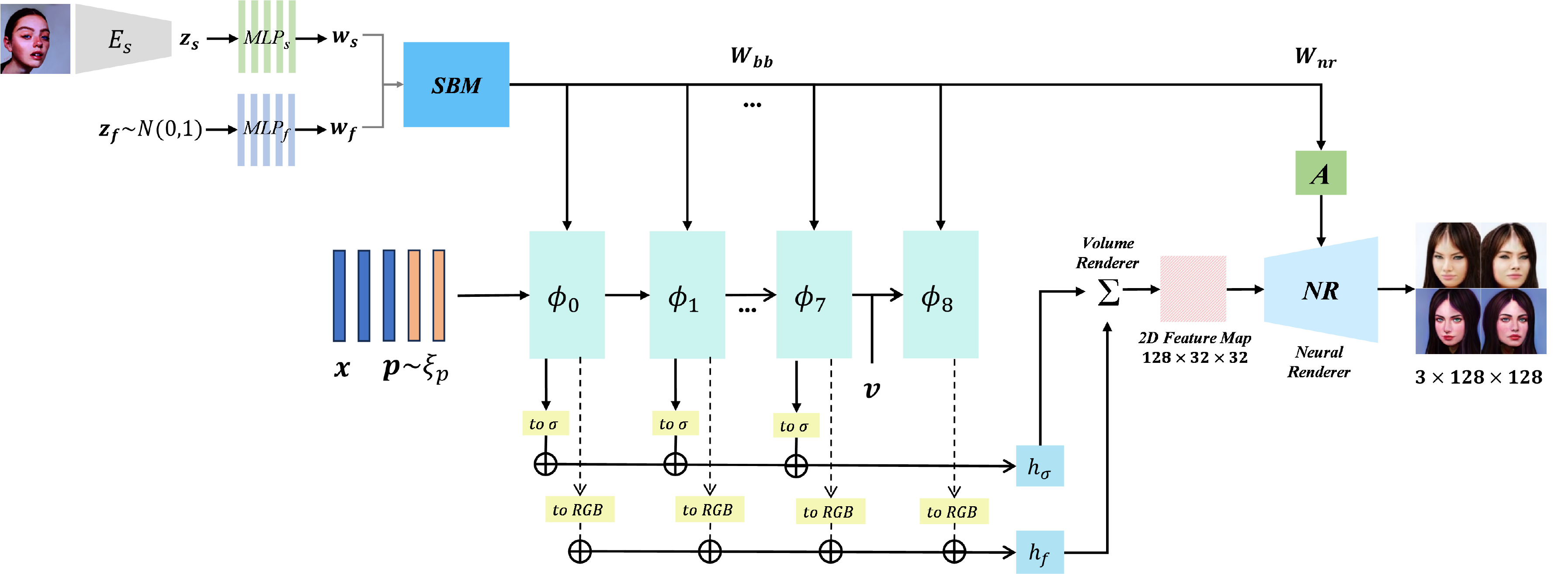}
\caption{\textbf{The pipeline of the generator in ArtNeRF.} Given an identity code $\boldsymbol{z_f}$ sampled from normal distribution and a style image $\boldsymbol{X_s}$, we first extract the style code using the style encoder $E_s$. Subsequently, dual mapping networks are utilized to map $\boldsymbol{z_f}, \boldsymbol{z_s}$ to $\boldsymbol{W_f}, \boldsymbol{W_s}$ in the $W^+$ space. The self-adaptive SBM module then blends $\boldsymbol{W_f}, \boldsymbol{W_s}$ based on a split index $i$ and injects the style information into the 3D generator. Given camera poses, real-time rendering of 3D-aware stylized faces can be achieved with the dense skip connections and the neural rendering module.} \label{fig:2}
\vspace{-10pt}
\end{figure}

\subsection{Preliminaries}\label{sec:3.1}
\textbf{Neural Radiance Fields.} A Neural Radiance Field (NeRF) implicitly represents the scene as a 5D function, enabling high-quality synthesis of novel views with multi-view consistency. Given a 3D point $x$, a radiance field $g_{\theta}$ is employed to map its position $(x,y,z)$ and the viewing direction $(\theta, \phi)$ to its RGB color $\boldsymbol{c}$ and volume density $\sigma$. To render a pixel, a ray $\boldsymbol{r}(t) = \boldsymbol{o} + t\boldsymbol{d}$ is cast from the camera origin $\boldsymbol{o}$ to the 3D space along the viewing direction $\boldsymbol{d}$, where $t\in [t_n, t_f]$ represents the distance from the sampling point to the camera origin. The color of the pixel can be rendered via volume rendering:

\begin{equation}
{\boldsymbol{C}}({\boldsymbol{r}}) = \int_{{t_n}}^{{t_f}} {T(t)\sigma ({\boldsymbol{r}}(t))} {\boldsymbol{c}}({\boldsymbol{r}}(t),{\boldsymbol{d}})dt,\;T(t) = {e^{ - \int_{{t_n}}^t {\sigma ({\boldsymbol{r}}(s))} ds}}
\end{equation}

\noindent where $T(t)$ is the cumulative transmittance from $t_n$ to $t$.

\subsection{Self-supervised Style Encoder}\label{sec:3.2}
Style encoder used to extract features is indispensable in style transfer tasks. However, utilizing an VGG-based encoder with randomly initialized parameters may lead to inconsistent styles and confusion. Moreover, since we need to generate 3D-aware images, it is crucial to prevent the leakage of pose information from the reference images into the style latent codes. To tackle this problem, we leverage a style encoder with strong expressive capability following~\cite{ref_proc8}.

\begin{figure}
\vspace{-10pt}
\includegraphics[width=\textwidth]{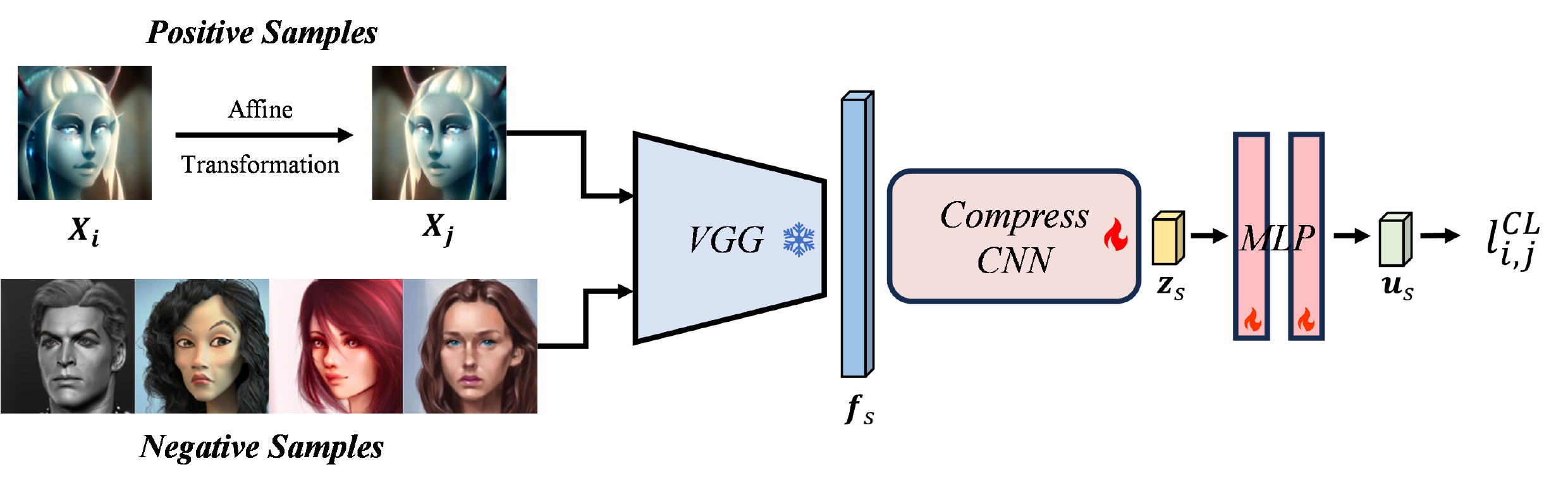}
\caption{The architecture of the style encoder $E_s$.} \label{fig:3}
\vspace{-5pt}
\end{figure}

The overall structure of the style encoder is illustrated in Fig. \ref{fig:3}. A pretrained VGG19 is utilized to extract style features $\boldsymbol{f_s}$ from input images, followed by a compress-CNN to reduce the dimension of $\boldsymbol{f_s}$ to 512. The 512-dim vectors serve as the style codes $\boldsymbol{z_s}$. To facilitate the contrastive learning process, a projection head is applied to further map $\boldsymbol{z_s}$ to their representation vectors $\boldsymbol{u_s}$. During training, each batch contains $2N$ images, where $\boldsymbol{X_i}, \boldsymbol{X_j}$ are positive samples ($\boldsymbol{X_i}$ is a style image and $\boldsymbol{X_j}$ is an augmented sample via affine transformation), and the remaining $2N-2$ images serve as negative samples. We use the following objective function to optimize the compress-CNN :

\begin{equation}
\ell _{i,j}^{CL} =  - \log \frac{{\exp ({\mathop{\rm sim}\nolimits} ({{\boldsymbol{u_i}}},{{\boldsymbol{u_j}}})/\tau )}}{{\sum\limits_{k = 1,k \ne i,k \ne j}^{2N} {\exp ({\mathop{\rm sim}\nolimits} ({{\boldsymbol{u_i}}},{{\boldsymbol{u_k}}})/\tau)} }}
\end{equation}

\noindent where $\rm sim(\cdot, \cdot)$ is the cosine similarity between embeddings, $\tau$ is the temperature coefficient, and $(\boldsymbol{u_i}, \boldsymbol{u_j})$ represents the contrastive learning representations for $(\boldsymbol{X_i}, \boldsymbol{X_j})$. After training, augmented samples of the same style image will have style codes rich in style semantics but devoid of spatial structure, since they are pushed closer in the embedding space and their original spatial features are neutralized.

\subsection{Conditional Generative Radiance Field}\label{sec:3.3}

pi-GAN~\cite{ref_proc11} is a NeRF-based 3D-aware face generation framework. We start from pi-GAN to design our generator considering its concise and effective backbone along with its relatively small training overhead. In this work, we extend and enhance pi-GAN to address the 3D-aware image stylization problem. Specifically, our improved neural generative radiance field comprises three main components: mapping network, style blending module (SBM), and conditional radiance field with dense skip connections.

\textbf{Mapping network and SBM Module.} 
Let $\boldsymbol{z_f}$ denote the identity latent code, and $\boldsymbol{z_s}$ represent the style code obtained from the style image. We first utilize two mapping networks with unshared parameters to respectively map $\boldsymbol{z}$ and $\boldsymbol{z_s}$ from $z$ space to $\boldsymbol{w}$ and $\boldsymbol{w_s}$ in $W^+$ space to achieve feature decoupling. \\
\indent To incorporate the two latent codes into our backbone, we design a style blending module SBM. Since different layers are responsible for learning facial semantic information at various levels, simply setting fixed blending weights for each layer is not advisable, which will inevitably cause mode collapse. Inspired by this, we first mix $\boldsymbol{w_f}$ and $\boldsymbol{w_s}$ using a learnable weight vector $\boldsymbol{\alpha}$, then feed the mixed code into our model to achieve style mixing of facial semantic information at multiple levels. The SBM module can be formulated as Eq.\ref{eq:3}. We omit the batch dimension for simplicity.

\begin{equation}
\begin{gathered}
\label{eq:3}
{\boldsymbol{W_{fused}} = \rm{Concat}(\boldsymbol{W_{bb}}; \boldsymbol{W_{nr}}}) \\
{\boldsymbol{W_{bb}}} = {\boldsymbol{\alpha}[:k]} \odot {{\boldsymbol{w_s}[:k, :]}} + (1 - {\boldsymbol{\alpha}[:k]}) \odot {{\boldsymbol{w_f}[:k, :]}} \\
{\boldsymbol{W_{nr}}} = {\boldsymbol{\alpha}[k:]} \odot \rm{Trans}({{\boldsymbol{w_s}[\mathnormal{k}:, :]}}) + (1 - {\boldsymbol{\alpha}[\mathnormal{k}:]}) \odot \rm{Trans}({{\boldsymbol{w_f}[\mathnormal{k}:, :]}})
\end{gathered}
\end{equation}

\noindent where $bb$ and $nr$ denote backbone and neural rendering, $\rm{Concat}(\cdot;\cdot)$ and $\odot$ denote channel-wise concatenation and element-wise multiplication, $[:]$ represent the slicing operation in PyTorch. In practice, $k$ is the number of layers in the backbone, $\boldsymbol{\alpha}$ is a learnable weight vector with a shape of $[n]$, $\boldsymbol{w_f}$ and $\boldsymbol{w_s}$ both have a shape of $[n, 256]$, where $n$ is the number of layers requiring style mixing. To flexibly adjust the degree of stylization, we introduce a split index $i$ in SBM. When $i$ is specified, $\boldsymbol{\alpha}[:i]=1$ and $\boldsymbol{\alpha}[i:]$ remains unchanged. This strategy ensures layers with indices less than $i$ only affected by $\boldsymbol{w_f}$, while layers with indices greater than $i$ influenced by both $\boldsymbol{w_f}$ and $\boldsymbol{w_s}$. Consequently, we can perform style blending on the backbone and the neural rendering module (Sec \ref{sec:3.4}). Note that directly injecting style vector into the neural rendering module may be improper as its feature space differs from the backbone. Hence, we apply projection units to $\boldsymbol{w_f}$ and $\boldsymbol{w_s}$ to refine them before the injection operation.

\textbf{Conditional Radiance Field with Dense Skip Connections.} The proposed conditional radiance field takes as input not only 3D positions $\boldsymbol{x}$ in the camera coordinate system and a camera pose $\boldsymbol{p}$ but also a fused conditioning latent code $\boldsymbol{W_{bb}}$. Therefore, properly injecting $\boldsymbol{W_{bb}}$ into the backbone is crucial for the generation performance. As is shown in Fig.\ref{fig:2}, the backbone consists of two parts: a sequence of $n$ FiLM layers and dense skip connections. The FiLM layer sequence can be formulated as follows:

\begin{equation}
\begin{gathered}
\Phi ({\boldsymbol{x}}) = {\phi _{n - 1}} \circ {\phi _{n - 2}} \circ ... \circ {\phi _0}({\boldsymbol{x}}) \\
{\phi _i}({{\boldsymbol{x_i}}}) = \sin ({{\boldsymbol{\gamma_i}}} \cdot ({{\boldsymbol{W_i}}}{{\boldsymbol{x_i}}} + {{\boldsymbol{b_i}}}) + {{\boldsymbol{\beta_i}}})
\end{gathered}
\end{equation}

\noindent where $\boldsymbol{x_i}$ is the input of the $i$-th FiLM layer, $\boldsymbol{W_i}, \boldsymbol{b_i}$ are learnable parameters and $\gamma_i, \beta_i$ are modulation coefficients projected from $\boldsymbol{W_{bb}}$. Note that $\phi_{n-1}$ also takes the viewing direction $\boldsymbol{v}$ as input to model view-dependant appearance, we omit it for brevity. To mitigate the ripple-like artifacts observed during training pi-GAN, we draw inspiration from StyleGAN2's~\cite{ref_proc18} improvements to StyleGAN. We discard the progressive growing training strategy used in pi-GAN and optimize the generator with a structure featuring dense skip connections. This modification ensures that feature maps from different layers can mutually contribute to the final output, thereby increasing the strength of gradient back-propagation and preventing training collapse. Optimized formulas for volume density and feature calculation is shown in Eq.\ref{eq:5}:

\begin{equation}
\begin{gathered}
\label{eq:5}
\sigma ({\boldsymbol{x}}) = {h_\sigma}(\sum\limits_{i = 0}^{n - 2} {{\lambda _i}({{\boldsymbol{M_i}}})}) \\
{\boldsymbol{f}}({\bf{x}}) = {h_f}(\sum\limits_{i = 0}^{n - 1} {{\mu _i}({{\boldsymbol{M_i}}})} )
\end{gathered}
\end{equation}

\noindent where ${{\boldsymbol{M_i}}}$ represents the output of $\phi_i$, $\lambda_i$ is the $i$-th volume density prediction layer and $\mu_i$ is the $i$-th feature prediction layer inspired by ~\cite{ref_proc12}. $h_{\sigma}, h_{f}$ are used to clamp the volume density $\sigma_i \in\mathbb{R}^1$ and the feature values $\boldsymbol{f_i}\in\mathbb{R}^{M_f}$. Let $\{{{\boldsymbol{x_i}}}\} _{i = 1}^{{N_s}}$ denote the $N_s$ sampling points along a ray, with volume density $\sigma_i$ and feature values $\boldsymbol{f_i}$ of each point, the volume rendering can be defined as follows:

\begin{equation}
{\pi_{vol}}:{(\mathbb{R}\times {\mathbb{R}^{{M_f}}})^{{N_s}}} \mapsto {\mathbb{R}^{{M_f}}},\;\;\{ {\sigma _i},{{\boldsymbol{f_i}}}\} _{i = 1}^{{N_s}} \mapsto {\boldsymbol{f}}
\end{equation}

By performing volume rendering to all rays, we can rapidly obtain the complete feature map ${\boldsymbol{F}} \in {\mathbb{R}^{{M_f}\times32\times32}}$ with relatively small GPU overhead. We can intuitively consider $\boldsymbol{F}$ as a texture representation of the final image.

\subsection{Neural Rendering Module}\label{sec:3.4}

\begin{figure}
\includegraphics[width=\textwidth]{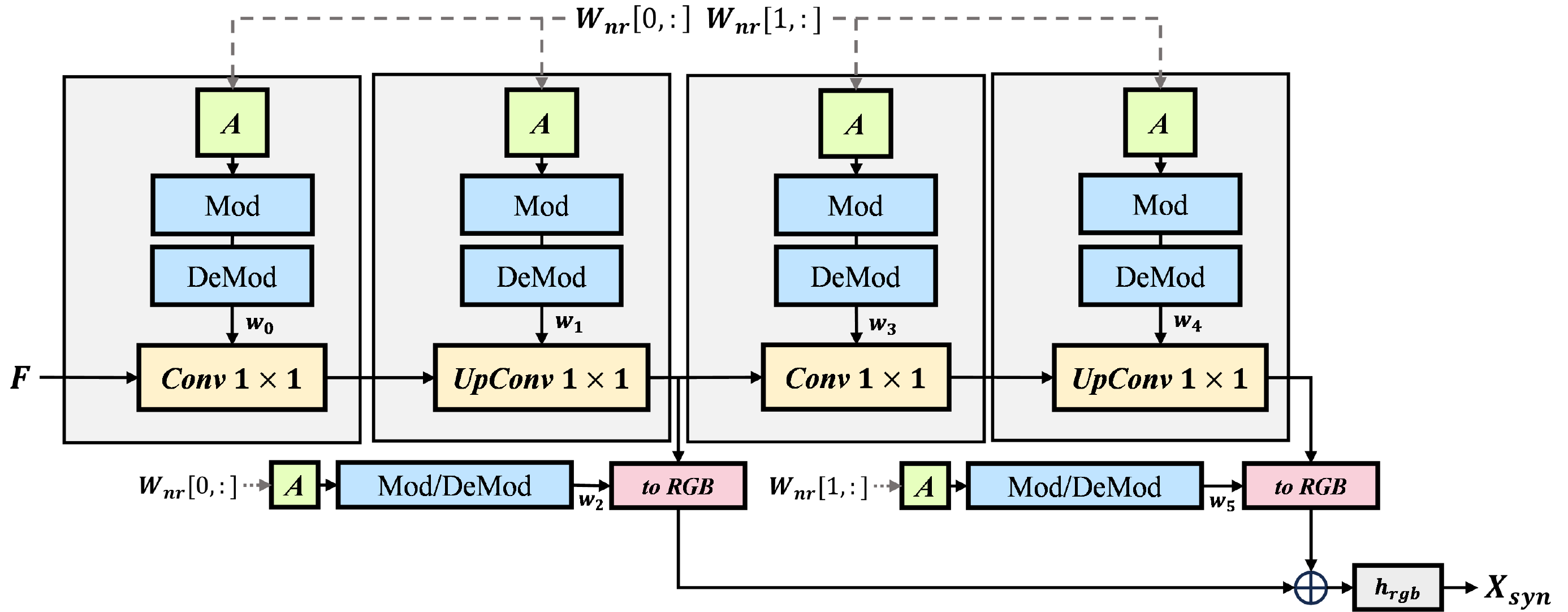}
\caption{The architecture of the neural rendering module.} \label{fig:4}
\vspace{-5pt}
\end{figure}

The integration of the neural rendering module can remarkably enhance the expressiveness of the generator, enabling the synthesis of high-quality images at higher resolutions with faster inference speed.
GIRAFFE~\cite{ref_proc12} first introduces a neural renderer composed of 2D CNNs into the model. However, EG3D~\cite{ref_proc15} indicates that overly deep 2D CNNs and excessive $3\times3$ convolution operations undermine the 3D consistency of the final results. Therefore, our neural rendering module is composed of shallow $1\times1$ convolutions. $1\times1$ convolutions enhance the network's ability to model information between channels of 2D feature maps and help avoid the fusion of local spatial contexture in feature maps, further ensuring multi-view consistency. The proposed neural rendering module leverages ModConv and upsampling layers in StyleGAN2, with a to\_rgb layer to facilitate the reuse of features between adjacent blocks. Introducing ModConv allows for the reuse of $\boldsymbol{W_{nr}}$ during super-resolution, which can refine the details of low-resolution results. Given a 2D low-resolution feature map $\boldsymbol{F}\in\mathbb{R}^{{M_f}\times32\times32}$, we can generate the final synthesized image $\boldsymbol{X_{syn}}\in\mathbb{R}^{3\times128\times128}$ following Fig. \ref{fig:4}, where the $A$ blocks are affine transformations applied to inject $\boldsymbol{W_{nr}}$ into the neural rendering module, $\mathbf{w_0}\dots\mathbf{w_5}$ are weight parameters for $1\times1$ convolutions and the $h_{rgb}$ is a clamp operation.

\subsection{Triple Discriminator Network}\label{sec:3.5}
We employ three discriminators to guide the generator to synthesize 3D-aware stylized images decently and appropriately. $D_r$ discriminates between fake natural faces and real natural faces, while $D_s$ discriminates between fake stylized faces and real ones. Together, they supervise the generator to ensure the generated images conforming to the distributions of the respective domains.
In order to further ensuring that the synthesized images match the style of the given reference images, we treat $\boldsymbol{w_s}$ as a sort of class labels inspired by~\cite{ref_proc19}. The task of generating stylized images can be naturally transformed into a cGAN problem. Therefore, we leverage a conditional discriminator $D_c$, which provides an additional supervision to the generator. Specifically, we apply an embedder head at the end of $D_c$. Let's denote the output of the global sum pooling layer in $D_c$ as $\boldsymbol{f_{gsp}}$. We first map a given style code to a feature embedding $\boldsymbol{f_{emb}}$ aligned with $\boldsymbol{f_{gsp}}$, then the dot product result of $\boldsymbol{f_{gsp}}$ and $\boldsymbol{f_{emb}}$ is added to the original output of $D_c$ to form the final output. The structure of $D_r$ and $D_s$ is similar to $D_c$, but without the embedder head. 

\subsection{Loss Functions}\label{sec:3.6}
Given a reference style image $\boldsymbol{X_s}$, we extract its style code $\boldsymbol{z_s}$ with $E_s$. We then sample an identity code $\boldsymbol{z_f}$ from a normal distribution and a camera pose $\boldsymbol{\xi}$ from a predefined distribution. On one side, we aim to synthesize fully stylized faces with the split index $i=0$ in SBM. On the other side, to ensure that generated stylized faces retain the original face identity during cross-domain adaption process, the generator should generate fully natural faces with $i=11$ in SBM. Additionally, a style consistency loss is leveraged to guarantee that the stylized faces share the same style as the reference images. During training, we maintain an embedding queue $Q$ that stores style codes from $i-1$-th batch. When we process $i$-th batch, we first sample a code $\boldsymbol{z_s^-}$ from $Q$ as a negative sample, we then instruct the generator to synthesize stylized face ${\boldsymbol{X_s^-}} = {G_{i = 0}}({{\boldsymbol{z_f}}},{{\boldsymbol{z_s^-}}})$. We feed $(\boldsymbol{X_s^-}, \boldsymbol{z_s^-})$ into $D_c$ as a pair of negative sample. The objective functions for $D_s, D_r, D_c$ can be formulated as follows:

\vspace{-10pt}
\begin{equation}
\begin{gathered}
{{\cal L}_{s}} = {\mathbb{E}_{{{\boldsymbol{z_f}}},{{\boldsymbol{X_s}}},{\boldsymbol{\xi}}}}\left[{f({D_s}({G_{i = 0}}({{\boldsymbol{z_f}}},\boldsymbol{z_s},{\boldsymbol{\xi}})))} \right] + {\mathbb{E}_{{{\boldsymbol{X_s}}}}}\left[{f(-{D_s}({{\boldsymbol{X_s}}})) + \lambda {{\left\| {\nabla {D_s}({{\boldsymbol{X_s}}})} \right\|}^2}}\right] \\
{{\cal L}_{r}} = {\mathbb{E}_{{{\boldsymbol{z_f}}},{\boldsymbol{\xi}}}} \left[ {f({D_r}({G_{i = 11}}({{\boldsymbol{z_f}}},{\boldsymbol{\xi}})))} \right] + {\mathbb{E}_{{{\boldsymbol{X_r}}}}} \left[ {f(-{D_r}({{\boldsymbol{X_r}}})) + \lambda {{\left\| {\nabla {D_r}({{\boldsymbol{X_r}}})} \right\|}^2}} \right] \\
{{\cal L}_{c}} = {\mathbb{E}_{{{\boldsymbol{z_f}}},{{\boldsymbol{X_s}}}}}\left[{f({D_c}({\boldsymbol{X_s^-}}, {\boldsymbol{z_s^-}})} \right] + {\mathbb{E}_{{{\boldsymbol{X_s}}}}}\left[{f(-{D_c}({{\boldsymbol{X_s}}},{E_s}({{\boldsymbol{X_s}}})))} \right]
\end{gathered}
\end{equation}

\noindent where $f(x) = - \log (1+e^{-x})$. We adopt the non-saturating GAN objective~\cite{ref_proc12} and $R_1$ gradient penalty to avoid mode collapse as well as stabilize the entire training process.

Finally, we need to ensure that all the generated faces are constrained within the same canonical space. To this end, the discriminator should predict the camera pose $\boldsymbol{\hat \xi} = (pitch, yaw)$ of the generated face and compute a pose consistency loss between $\boldsymbol{\hat \xi}$ and the previously sampled pose $\boldsymbol{\xi}$. We apply pose consistency loss for both natural faces (denoted as real) and stylized faces (denoted as style):

\begin{equation}
\begin{gathered}
{{\cal L}_{real-pose}} = \mathbb{E}{_\xi}{\left\| {{{{\boldsymbol{\hat \xi}_{real}}}} - {{\boldsymbol{\xi}_{real}}}} \right\|^2} \\
{{\cal L}_{style-pose}} = \mathbb{E}{_\xi}{\left\| {{{{\boldsymbol{\hat \xi}_{style}}}} - {{\boldsymbol{\xi}_{style}}}} \right\|^2}
\end{gathered}
\end{equation}

Given $\lambda_1, \lambda_2, \lambda_3$, which are weights to balance these objective functions, the entire training loss of ArtNeRF is:
\begin{equation}
\begin{gathered}
{{\cal L}_D} = {\lambda _1}({{\cal L}_{real}} + {{\cal L}_{real - pose}}) + {\lambda _2}({{\cal L}_{style}} + {{\cal L}_{style - pose}}) + {\lambda _3}{{\cal L}_{style - latent}} \\
{{\cal L}_G} =  - {{\cal L}_D}^{no-R1}
\end{gathered}
\end{equation}
where ${{\cal L}_D}^{no-R1}$ represents ${{\cal L}_D}$ without $R_1$ penalty term.

\section{Experiments}

\begin{figure}
\includegraphics[width=\textwidth]{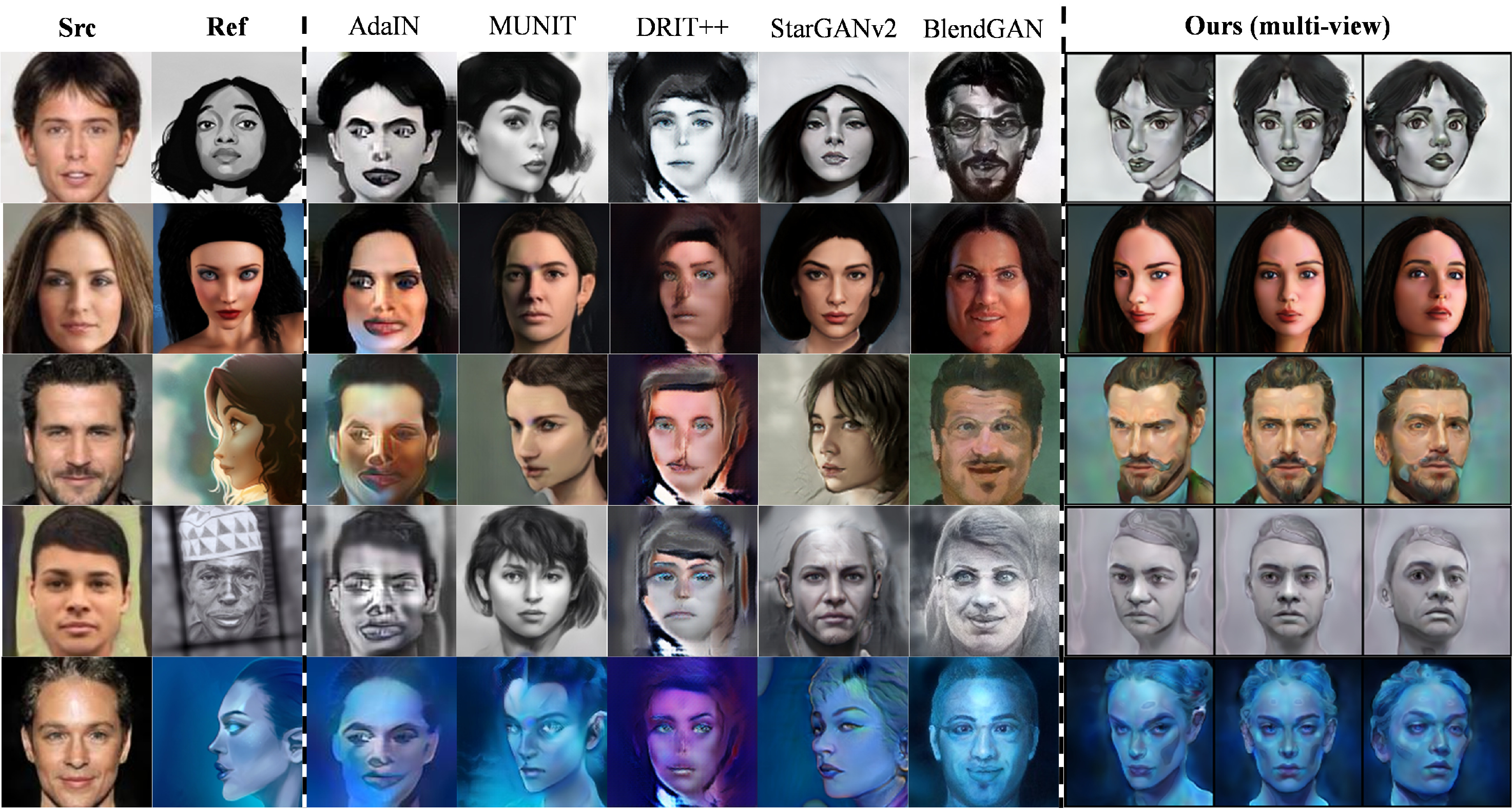}
\caption{Qualitative comparison of style-guided face synthesis between AdaIN~\cite{ref_proc21}, MUNIT~\cite{ref_proc1}, FUNIT~\cite{ref_proc2}, DRIT++~\cite{ref_proc3}, StarGANv2~\cite{ref_proc4}, BlendGAN~\cite{ref_proc8} and ours. Note that our model can not only generate reasonable stylized faces, but also produces 3D-aware results with high multi-view consistency.} \label{fig:6}
\vspace{-5pt}
\end{figure}

\textbf{Datasets.} We utilize CelebA~\cite{ref_proc20}, containing approximately 200k faces, as our source domain dataset. For the style domain, we employ AAHQ~\cite{ref_proc8}, an artistic dataset comprising around 24k high-quality stylized faces. All the images from the two datasets have been cropped and aligned properly, with a resolution of $128\times128$.

\textbf{Implementation details.} We use a two-stage strategy to train our model: base model pre-training and fine-tuning within the style domain. We train 200k steps for stage1 merely with $D_r$ using CelebA and 30k steps for stage2 with triple-branch discriminator module using both the two datasets. After stage1, our model can sufficiently learn prior knowledge about the distribution of natural faces. In stage 2, with the assistance of SBM module and triple-branch discriminator module, the base model will be decently guided to generate stylized faces in a cross-domain adaption manner. Our method is implemented using PyTorch and trained on single NVIDIA RTX 2080Ti GPU for about 3 days.

\subsection{Comparisons}

\textbf{Qualitative results.} Fig. \ref{fig:1} displays synthesized multi-view natural faces and their corresponding stylized faces at different levels ($i=0$ and $i=3$) within the SBM module. At $i=0$, the results exhibit the highest degree of stylization but lose some identity features. At $i=3$, a balance between stylization and identity preservation is achieved. Fig. \ref{fig:6} provides a qualitative comparison with baselines (using $i=3$). DRIT++ fails to learn the style consistency between generated faces and reference images, while AdaIN struggles with identity preservation. Although MUNIT and StarGANv2 produce reasonable results, they tend to overly inherit face poses from reference images. BlendGAN performs well in identity preservation and style consistency but lacks 3D-awareness. Our method excels in synthesizing 3D-aware images with robust multi-view consistency, achieving strong identity preservation and style consistency simultaneously.

\begin{table}
\centering
\caption{A thorough comparison of the functionality among several prevailing face cartoonization or stylization methods.}
\label{tab:1}
\begin{tabular}{l|l|l|l|l}
\hline
Method & Year & Reference-guided\hspace{2mm} & Arbitrary style\hspace{2mm} & 3D-aware\hspace{2mm}\\
\hline
CartoonGAN~\cite{ref_proc6} & CVPR18\hspace{2mm} & × & × & × \\
AniGAN~\cite{ref_proc22} & TMM21\hspace{2mm} & \checkmark & × & × \\
AdaIN~\cite{ref_proc21} & ICCV17\hspace{2mm} & \checkmark & \checkmark & × \\
MUNIT\cite{ref_proc1} & ECCV18\hspace{2mm} & \checkmark & \checkmark & × \\
FUNIT\cite{ref_proc2} & ICCV19\hspace{2mm} & \checkmark & \checkmark & × \\
DRIT++\cite{ref_proc3} & IJCV20\hspace{2mm} & \checkmark & \checkmark & × \\
StarGANv2~\cite{ref_proc4} & CVPR20\hspace{2mm} & \checkmark & \checkmark & × \\
BlendGAN~\cite{ref_proc8} & NIPS21\hspace{2mm} & \checkmark & \checkmark & × \\
JoJoGAN~\cite{ref_proc23} & ECCV22\hspace{2mm} & \checkmark & × & × \\
DualStyleGAN~\cite{ref_proc9} & CVPR22\hspace{2mm} & \checkmark & × & × \\
3DAvatarGAN~\cite{ref_proc16} & CVPR23\hspace{2mm} & \checkmark & × & \checkmark \\
DeformToon3D~\cite{ref_proc17} & ICCV23\hspace{2mm} & \checkmark & × & \checkmark \\
\textbf{ArtNeRF(Ours)} & \textbf{2024}\hspace{2mm} & $\boldsymbol{\checkmark}$ & $\boldsymbol{\checkmark}$ & $\boldsymbol{\checkmark}$ \\
\hline
\end{tabular}
\end{table}

\textbf{Quantitative results.} Table \ref{tab:1} demonstrates our method's capability to generate reference-guided 3D-aware faces with arbitrary styles, a task not addressed by existing methods. We provide quantitative comparisons against reference-guided image synthesis baselines using FID, KID, and IS metrics on 20k generated stylized images and 20k style reference images. Besides, we assess image diversity using LPIPS. Given a specified identity code, we select 10 reference styles randomly and generate 10 stylized faces, we then evaluate the LPIPS scores between every 2 results. This process is repeated for 1000 identity codes and the average of all scores constitute the final LPIPS score.BlendGAN differs from ArtNeRF in training settings and their training code is unavailable, so we reproduce the latent-guided (an identity code and a style code is sampled) results of BlendGAN and downsample them to $128\times128$ for comparison.

\begin{table}
\centering
\caption{Quantitative evaluation of style-guided face synthesis. We compare with methods that support face stylization with arbitrary reference styles.}
\label{tab:2}
\begin{tabular}{l|l|l|l|l}
\hline
Method & FID↓ & KID↓ & IS↑ & LPIPS↑ \\
\hline
AdaIN & 86.87 & 0.084 & 2.14 & 0.237 \\
MUNIT & 56.99 & 0.046 & 1.98 & 0.241 \\
DRIT++ & 89.79 & 0.069 & 2.02 & 0.231 \\
StarGANv2\hspace{2mm} & 34.24 & 0.022 & 2.50 & 0.389 \\
BlendGAN & 39.45 & 0.037 & \textbf{2.98} & 0.239 \\
Ours(i=3) & 13.80 & 0.0066 & 2.89 & 0.377 \\
Ours(i=0) & \textbf{12.09}\hspace{2mm} & \textbf{0.0052}\hspace{2mm} & 2.96\hspace{2mm} & \textbf{0.403}\hspace{2mm} \\
\hline
\end{tabular}
\end{table}

Tab \ref{tab:2} highlights our method's significant improvements over AdaIN, MUNIT, DRIT++, and StarGANv2 across quantitative metrics. Notably,  we use multi-view stylized faces for evaluation while other methods use fix-pose faces. It manifests the 3D-aware faces generated by our method possess higher visual quality and diversity than the baseline methods. BlendGAN is the SOTA in 2D reference-guided image synthesis with arbitrary style and our method exhibits slightly inferior IS compared to BlendGAN, suggesting that there is still room for our method to narrow the visual quality gap between 2D and 3D-aware methods.

\subsection{Ablation Study}
In this section, we conduct extensive ablation studies to assess the impact of various modules on the model's generative capability and demonstrate their effectiveness. Experiments involving the base model (stage 1) are conducted using the CelebA dataset, whereas experiments focusing on the final stylization model (stage 2) are carried out using the AAHQ dataset.

\begin{table}
\centering
\caption{We improve the generation capability of the base model in a progressive way. PG, DSC and NR denote progressive growing training strategy, dense skip connections and neural rendering, respectively.}
\label{tab:3}
\renewcommand{\arraystretch}{1.2}
\begin{tabular}{l|l|l|l|l}
\hline
 & FID↓ (10k) & FID↓ (20k)  & FID↓ (40k) & FID↓ (100k) \\
\hline
base model & 58.76 & 40.83 & 24.47 & 36.49 \\
\hline
-PG & 63.9 & 49.05 & 34.95 & 27.46 \\
\hline
-PG, +DSC & 61.59 & 40.39 & 24.61 & 17.28 \\
\hline
-PG, +DSC, +NR (ours) & \textbf{53.97} & \textbf{33.97} & \textbf{22.29} & \textbf{14.42} \\
\hline
\end{tabular}
\end{table}

\textbf{Generator network.} In Sec \ref{sec:3.3}, we enhance the pi-GAN baseline by omitting the progressive growing strategy, integrating dense skip connections into the backbone, and introducing a neural rendering module. These modifications lead to a significant enhancement in synthesized image quality. We assess their effectiveness by progressively incorporating them into the baseline model in stage 1 and comparing results in Tab \ref{tab:3}. Omitting the progressive growing strategy initially yields a slightly higher FID at the start of training but significantly lowers it towards the end. Adding dense skip connections further reduces the final FID. Finally, if the neural rendering module is applied, the visual quality of our results will be further refined during the entire training process.

\begin{table}
\centering
\caption{The rendering speed (fps) with (w/) and without (w/o) neural rendering module under different (res, ns) pairs. OOM denotes CUDA out of memory error.}
\label{tab:4}
\renewcommand{\arraystretch}{1.2}
\begin{tabular}{l|l|l|l|l|l|l|l|l|}
\hline
 & \multicolumn{2}{c|}{ns=8} & \multicolumn{2}{c|}{ns=16} & \multicolumn{2}{c|}{ns=32} & \multicolumn{2}{c|}{ns=48} \\
\hline
 & w/o nr\hspace{2mm} & w/ nr\hspace{2mm} & w/o nr\hspace{2mm} & w/ nr\hspace{2mm} & w/o nr\hspace{2mm} & w/ nr\hspace{2mm} & w/o nr\hspace{2mm} & w/ nr\hspace{2mm} \\
\hline
res=64 & 33.45 & \textbf{49.69} & 41.20 & \textbf{56.44} & 26.33 & \textbf{48.80} & 18.26 & \textbf{48.76}\\
\hline
res=128 & 25.57 & \textbf{53.60} & 14.11 & \textbf{52.40} & 7.29 & \textbf{42.69} & OOM & \textbf{38.54}\\
\hline
res=256 & 7.22 & \textbf{43.40} & 3.25 & \textbf{32.93} & OOM & \textbf{21.93} & OOM & \textbf{15.58}\\
\hline
\end{tabular}
\end{table}

\textbf{Neural rendering.} The neural rendering module enhances the low-resolution texture representation from the backbone, producing the final high-resolution face image and substantially accelerating the inference process. We analyze how neural rendering impacts inference speed across different resolutions (res) and samples per ray (ns). We conduct experiments with our stylization model. As detailed in Tab \ref{tab:4}, our findings demonstrate a significant enhancement in inference speed across nearly all $(\rm{res}, \rm{ns})$ pairs. Notably, when $\rm{res}=256, \rm{ns}=16$, neural rendering improves inference speed by \textbf{$10\times$} compared to the original structure. This capability allows the model to efficiently handle diverse $(\rm{res}, \rm{ns})$ settings, facilitating high-quality real-time rendering essential for VR/AR applications.

\vspace{-5pt}
\section{Conclusion}
In this paper, we propose a novel 3D-aware image stylization method ArtNeRF, enabling the generation of faces with arbitrary styles. We achieve this goal by enhancing a NeRF-GAN baseline with dense skip connections and a neural rendering module, proposing an SBM module to integrate style control vectors into the generator and leveraging a triple-branch discriminator to improve style and multi-view consistency. Extensive experiments illustrate the effectiveness of ArtNeRF. However, our model still has limitations. Although reasonable faces can be generated with most camera viewpoints, our model cannot tackle with extreme views or synthesize 360° images of human heads due to dataset constraints. Future work will incorporate advanced 3D representations like 3D Gaussian Splatting~\cite{ref_proc24} to further enhance image quality and rendering speed.

\bibliographystyle{splncs04}
\bibliography{reference}

\begin{thebibliography}{10}
\providecommand{\url}[1]{\texttt{#1}}
\providecommand{\urlprefix}{URL }
\providecommand{\doi}[1]{https://doi.org/#1}

\bibitem{ref_proc16}
Abdal, R., Lee, H.Y., Zhu, P., Chai, M., Siarohin, A., Wonka, P., Tulyakov, S.: 3davatargan: Bridging domains for personalized editable avatars. In: Proceedings of the IEEE/CVF Conference on Computer Vision and Pattern Recognition. pp. 4552--4562 (2023)

\bibitem{ref_proc15}
Chan, E.R., Lin, C.Z., Chan, M.A., Nagano, K., Pan, B., De~Mello, S., Gallo, O., Guibas, L.J., Tremblay, J., Khamis, S., et~al.: Efficient geometry-aware 3d generative adversarial networks. In: Proceedings of the IEEE/CVF conference on computer vision and pattern recognition. pp. 16123--16133 (2022)

\bibitem{ref_proc11}
Chan, E.R., Monteiro, M., Kellnhofer, P., Wu, J., Wetzstein, G.: pi-gan: Periodic implicit generative adversarial networks for 3d-aware image synthesis. In: Proceedings of the IEEE/CVF conference on computer vision and pattern recognition. pp. 5799--5809 (2021)

\bibitem{ref_proc6}
Chen, Y., Lai, Y.K., Liu, Y.J.: Cartoongan: Generative adversarial networks for photo cartoonization. In: Proceedings of the IEEE conference on computer vision and pattern recognition. pp. 9465--9474 (2018)

\bibitem{ref_proc4}
Choi, Y., Uh, Y., Yoo, J., Ha, J.W.: Stargan v2: Diverse image synthesis for multiple domains. In: Proceedings of the IEEE/CVF conference on computer vision and pattern recognition. pp. 8188--8197 (2020)

\bibitem{ref_proc23}
Chong, M.J., Forsyth, D.: Jojogan: One shot face stylization. In: European Conference on Computer Vision. pp. 128--152. Springer (2022)

\bibitem{ref_proc14}
Deng, Y., Yang, J., Xiang, J., Tong, X.: Gram: Generative radiance manifolds for 3d-aware image generation. In: Proceedings of the IEEE/CVF Conference on Computer Vision and Pattern Recognition. pp. 10673--10683 (2022)

\bibitem{ref_proc5}
Goodfellow, I., Pouget-Abadie, J., Mirza, M., Xu, B., Warde-Farley, D., Ozair, S., Courville, A., Bengio, Y.: Generative adversarial nets. Advances in neural information processing systems  \textbf{27} (2014)

\bibitem{ref_proc7}
He, B., Gao, F., Ma, D., Shi, B., Duan, L.Y.: Chipgan: A generative adversarial network for chinese ink wash painting style transfer. In: Proceedings of the 26th ACM international conference on Multimedia. pp. 1172--1180 (2018)

\bibitem{ref_proc21}
Huang, X., Belongie, S.: Arbitrary style transfer in real-time with adaptive instance normalization. In: Proceedings of the IEEE international conference on computer vision. pp. 1501--1510 (2017)

\bibitem{ref_proc1}
Huang, X., Liu, M.Y., Belongie, S., Kautz, J.: Multimodal unsupervised image-to-image translation. In: Proceedings of the European conference on computer vision (ECCV). pp. 172--189 (2018)

\bibitem{ref_proc18}
Karras, T., Laine, S., Aittala, M., Hellsten, J., Lehtinen, J., Aila, T.: Analyzing and improving the image quality of stylegan. In: Proceedings of the IEEE/CVF conference on computer vision and pattern recognition. pp. 8110--8119 (2020)

\bibitem{ref_proc24}
Kerbl, B., Kopanas, G., Leimk{\"u}hler, T., Drettakis, G.: 3d gaussian splatting for real-time radiance field rendering. ACM Transactions on Graphics  \textbf{42}(4),  1--14 (2023)

\bibitem{ref_proc3}
Lee, H.Y., Tseng, H.Y., Huang, J.B., Singh, M., Yang, M.H.: Diverse image-to-image translation via disentangled representations. In: Proceedings of the European conference on computer vision (ECCV). pp. 35--51 (2018)

\bibitem{ref_proc22}
Li, B., Zhu, Y., Wang, Y., Lin, C.W., Ghanem, B., Shen, L.: Anigan: Style-guided generative adversarial networks for unsupervised anime face generation. IEEE Transactions on Multimedia  \textbf{24},  4077--4091 (2021)

\bibitem{ref_proc2}
Liu, M.Y., Huang, X., Mallya, A., Karras, T., Aila, T., Lehtinen, J., Kautz, J.: Few-shot unsupervised image-to-image translation. In: Proceedings of the IEEE/CVF international conference on computer vision. pp. 10551--10560 (2019)

\bibitem{ref_proc8}
Liu, M., Li, Q., Qin, Z., Zhang, G., Wan, P., Zheng, W.: Blendgan: Implicitly gan blending for arbitrary stylized face generation. Advances in Neural Information Processing Systems  \textbf{34},  29710--29722 (2021)

\bibitem{ref_proc20}
Liu, Z., Luo, P., Wang, X., Tang, X.: Deep learning face attributes in the wild. In: Proceedings of the IEEE international conference on computer vision. pp. 3730--3738 (2015)

\bibitem{ref_proc13}
Mildenhall, B., Srinivasan, P.P., Tancik, M., Barron, J.T., Ramamoorthi, R., Ng, R.: Nerf: Representing scenes as neural radiance fields for view synthesis. Communications of the ACM  \textbf{65}(1),  99--106 (2021)

\bibitem{ref_proc19}
Miyato, T., Koyama, M.: cgans with projection discriminator. arXiv preprint arXiv:1802.05637  (2018)

\bibitem{ref_proc12}
Niemeyer, M., Geiger, A.: Giraffe: Representing scenes as compositional generative neural feature fields. In: Proceedings of the IEEE/CVF Conference on Computer Vision and Pattern Recognition. pp. 11453--11464 (2021)

\bibitem{ref_proc10}
Schwarz, K., Liao, Y., Niemeyer, M., Geiger, A.: Graf: Generative radiance fields for 3d-aware image synthesis. Advances in Neural Information Processing Systems  \textbf{33},  20154--20166 (2020)

\bibitem{ref_proc9}
Yang, S., Jiang, L., Liu, Z., Loy, C.C.: Pastiche master: Exemplar-based high-resolution portrait style transfer. In: Proceedings of the IEEE/CVF conference on Computer Vision and Pattern Recognition. pp. 7693--7702 (2022)

\bibitem{ref_proc17}
Zhang, J., Lan, Y., Yang, S., Hong, F., Wang, Q., Yeo, C.K., Liu, Z., Loy, C.C.: Deformtoon3d: Deformable neural radiance fields for 3d toonification. In: Proceedings of the IEEE/CVF International Conference on Computer Vision. pp. 9144--9154 (2023)

\end{thebibliography}

\end{document}